\documentclass[runningheads]{llncs}

\usepackage[T1]{fontenc}
\usepackage{graphicx}
\usepackage{svg}
\usepackage{booktabs}
\usepackage{amsmath,amssymb}
\usepackage{multirow}
\usepackage{array}
\usepackage{hyperref}
\usepackage{caption}
\usepackage{subcaption}

\begin{document}

\title{GraM‑Diff: A Unified Graph–Mamba Diffusion Framework for EEG‑Based Alzheimer’s Disease Data Generation and Diagnosis}
\titlerunning{GraM-Diff}

\author{
M. Tanveer\inst{1} \and
Ayush Singh Rana\inst{1}$^{\dagger}$ \and
Sanskriti Jain\inst{1}$^{\dagger}$ \and
Arnav Kumar\inst{1}$^{\dagger}$ \and\\
Aryaman Tiwari\inst{1}$^{\dagger}$ \and
A. Rahaman\inst{1}$^{*}$ \and
A. Quadir\inst{1}$^{*}$ \and
M. Sajid\inst{1}$^{*}$
}


\authorrunning{M. Tanveer et al.}

\institute{
Indian Institute of Technology Indore, Indore 453552, Madhya Pradesh, India\\
\email{\{mtanveer,~cse240001015,~cse240001064,~cse240001013,~cse240001014,\\phd2401141001,~mscphd2207141002,~phd2101241003\}@iiti.ac.in}
}

\maketitle

\begingroup
\renewcommand{\thefootnote}{}
\footnotetext{$^{\dagger}$ Ayush Singh Rana, Sanskriti Jain, Arnav Kumar, and Aryaman Tiwari contributed equally to this work.}
\footnotetext{$^{*}$ A. Rahaman, A. Quadir, and M. Sajid contributed equally to this work.}
\endgroup

\begin{abstract}
Electroencephalography (EEG) is a promising, non-invasive, and cost-effective modality for Alzheimer’s disease (AD) detection, but deep learning methods are limited by small and imbalanced clinical datasets. Generative augmentation offers a solution, yet existing approaches rely on inefficient class-specific models or fail to capture complex spatial and temporal brain dynamics. To address this, we propose GraM‑Diff, a unified classifier-guided Graph--Mamba diffusion framework for EEG synthesis. It embeds Graph Convolutional Networks within a diffusion U-Net to model inter-electrode connectivity and Bidirectional Mamba state-space blocks for linear-complexity long-range temporal modeling. Latent-space classifier guidance lets a single model generate both healthy and pathological EEG within a shared representation, avoiding fragmented per-cohort pipelines. Across four EEG-based AD benchmarks, synthetic augmentation improves classification, yields superior Context-FID and correlation scores over strong generative baselines, and enhances robustness in data-scarce settings. The source code of the proposed method is at \url{https://github.com/mtanveer1/GraM-Diff}.
\end{abstract}

\section{Introduction}
Alzheimer's Disease (AD) is a progressive neurodegenerative disorder and the leading cause of dementia worldwide, marked by amyloid-beta plaques and hyperphosphorylated tau tangles that drive synaptic dysfunction~\cite{tanveer2024fuzzy}. Roughly 7.2 million Americans aged 65+ live with AD, projected to reach 13.8 million by 2060~\cite{AlzheimerAssociation2025}. Because AD has a long preclinical phase and approved disease-modifying therapies are most effective early, scalable early-diagnostic strategies are critical. Electroencephalography (EEG) is a non-invasive, cost-effective modality for detecting cognitive-decline biomarkers, with accessibility and high temporal resolution well-suited to longitudinal assessment. Computational EEG analysis has progressed from handcrafted spectral/complexity features with classical classifiers~\cite{tanveer2024fuzzy} to deep CNN~\cite{alghamdi2025novel}, GNN~\cite{klepl2022eeg}, and Transformer~\cite{wang2024adformer} architectures; Denoising Diffusion Probabilistic Models (DDPMs) have recently emerged for EEG synthesis under limited data~\cite{vetter2024generating}.

Despite this progress, annotated EEG remains scarce. Existing generative approaches often use \emph{separate} models for healthy and pathological cohorts, increasing overhead and reducing flexibility. Public AD-EEG corpora are small: even combining AD-Auditory, ADSZ, APAVA, and BrainLat yields limited subject counts for high-capacity training~\cite{wang2025lead}, while class imbalance and privacy-driven data silos (GDPR, HIPAA) further hinder generalization~\cite{umair2025privacy}.

To address these challenges, we propose GraM-Diff, a unified Graph--Mamba diffusion framework that combines topology-aware spatial modeling with long-range temporal modeling in a single generative architecture. Unlike perturbation-based augmentation, it directly generates multichannel EEG conditioned on diagnostic state while preserving spatio-temporal structure relevant to downstream classification~\cite{sharma2023medic,carrle2023generation}. Our key contributions are: \textbf{(i)} a \emph{unified classifier-guided DDPM} that conditions a single generator on diagnostic state via latent-space guidance, enabling scalable class-conditioned multichannel EEG generation within a shared representation rather than per-cohort generators; \textbf{(ii)} \emph{topology-aware spatial modeling}, integrating graph-based spatial priors directly into a diffusion U-Net to capture non-Euclidean inter-electrode relationships~\cite{klepl2022eeg}; \textbf{(iii)} \emph{linear-complexity temporal modeling} via Bidirectional Mamba state-space blocks with $\mathcal{O}(L)$ cost over extended recordings~\cite{gui2024eegmamba}; and \textbf{(iv)} \emph{noise-robust LogRoPE guidance}, using patch-embedding conditioning with Logarithmic Rotary Positional Embeddings for stable representation learning under diffusion noise~\cite{wang2024adformer}.

\section{Related Work}
\label{sec:related}
\textbf{Discriminative EEG-AD models.} AD diagnosis has evolved from statistical modeling to deep architectures~\cite{tanveer2024fuzzy}, including 3D CNNs on spatio-temporal-spectral tensors~\cite{alghamdi2025novel} and ensembles over raw EEG. These typically treat channels as independent sequences or grid images, ignoring the explicit graph topology of brain connectivity~\cite{klepl2022eeg}. Multi-granularity Transformers model multiple time scales for EEG-AD assessment~\cite{wang2024adformer} but incur quadratic complexity on long recordings and have not been used generatively.

\textbf{Data scarcity and generative augmentation.} Aggregated public AD-EEG corpora contain few subjects (often $<330$), making high-capacity training prone to overfitting~\cite{wang2025lead}. GAN-based augmentation can improve accuracy~\cite{carrle2023generation} but suffers training instability and mode collapse and usually synthesizes features rather than raw multichannel signals. Diffusion models offer stabler training and avoid mode collapse~\cite{vetter2024generating}, and have been applied to EEG super-resolution and psychiatric classification~\cite{sharma2023medic}. Yet existing EEG diffusion models are generally not conditioned on AD pathology and lack classifier guidance to retain disease-specific signatures. 

Recurrent neural networks and Transformers often struggle with long-sequence EEG modelling, either due to vanishing-gradient issues or the quadratic computational cost of self-attention. Mamba state-space models address these limitations through linear-time sequence modelling, making them suitable for efficient long-range temporal representation learning~\cite{gui2024eegmamba}. In GraM-Diff, we incorporate a bidirectional Mamba block within the diffusion bottleneck to capture both causal and anti-causal dependencies across extended EEG recordings. Overall, GraM-Diff combines topology-aware spatial priors, linear-time bidirectional Mamba modelling, and LogRoPE-based multi-granular conditioning within a unified generative--discriminative framework.

\section{Proposed Methodology}
\label{sec:method}
\begin{figure}
    \centering
    \includegraphics[width=1\linewidth]{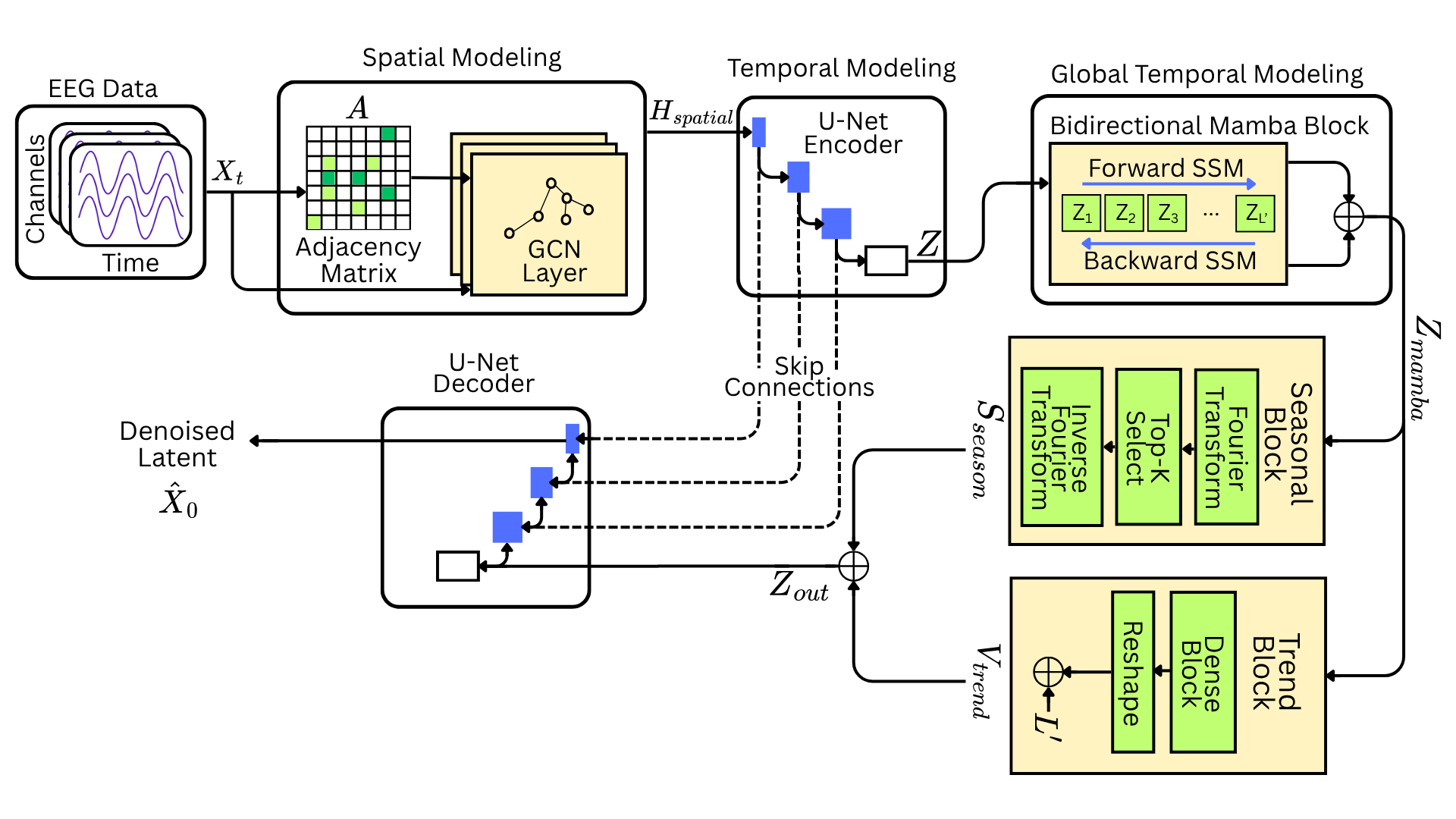}
    \caption{Architecture of GraM-Diff. EEG data are first passed through a
    GCN to obtain spatially enriched features. The U-Net encoder extracts
    hierarchical temporal representations, which enter a bidirectional Mamba
    bottleneck. The decoder reconstructs the denoised latent and decomposes
    it into trend and seasonal components. A time-conditioned transformer
    classifier operating on latent features provides gradients to guide the
    diffusion process.}
    \label{fig:architecture}
\end{figure}

As shown in Fig.~\ref{fig:architecture}, GraM-Diff combines spatial electrode relationships and long-range temporal modeling within a single architecture, together with a latent-space guidance mechanism, to generate Alzheimer's stage-conditioned EEG.

\subsection{Problem Formulation and Diffusion Process}
Let the clean multichannel EEG signal be $\mathbf{X}_0=[x_1,\dots,x_L]\in\mathbb{R}^{B\times C\times L}$, where $B$ is the batch size, $C$ the number of channels, and $L$ the temporal length. Following the DDPM paradigm to model $q(\mathbf{X}_0)$, the denoising network operates on latent features $\mathbf{h}^t\in\mathbb{R}^{B\times C\times L\times D}$ extracted by the U-Net encoder at timestep $t$. The forward process corrupts $\mathbf{X}_0$ into $\mathbf{X}_t$ over $t\in\{1,\dots,T\}$ ($T{=}1000$) as a Gaussian Markov chain:
\begin{equation}
q(\mathbf{X}_t\mid\mathbf{X}_0)=\mathcal{N}\!\left(\mathbf{X}_t;\sqrt{\bar{\alpha}_t}\,\mathbf{X}_0,(1-\bar{\alpha}_t)\mathbf{I}\right),\qquad \bar{\alpha}_t=\textstyle\prod_{i=1}^{t}(1-\beta_i),
\end{equation}
with $\beta_i\in(0,1)$. Because linear schedules abruptly destroy fine-grained high-frequency EEG structure, we adopt a cosine schedule $\bar{\alpha}_t=f(t)/f(0)$, $f(t)=\cos^2\!\big(\frac{t/T+s}{1+s}\cdot\frac{\pi}{2}\big)$ with $s{=}0.008$. The network is trained to predict the clean signal $\hat{\mathbf{X}}_0$ rather than the injected noise; the noise term is recovered analytically as $\epsilon_\theta(\mathbf{h}^t,t)=(\mathbf{X}_t-\sqrt{\bar{\alpha}_t}\hat{\mathbf{X}}_0)/\sqrt{1-\bar{\alpha}_t}$ and used in classifier-guided sampling. Diffusion and guidance operate on sequence-first tensors $\mathbb{R}^{B\times L\times C}$, internally permuted to channel-first $\mathbb{R}^{B\times C\times L}$ for U-Net and GCN processing.

\subsection{Stage 1: Topology-Aware Spatial Encoding (GCN)}
Standard convolutions treat EEG channels as grid-adjacent rows, ignoring scalp geometry. We instead pass $\mathbf{X}_t$ through a GCN with electrodes as nodes. Using 3D coordinates $\mathbf{c}_i\in\mathbb{R}^3$ from the International 10--20 system, a geometry-informed adjacency $\mathbf{A}\in\mathbb{R}^{C\times C}$ is built with a distance-based Gaussian kernel,
\begin{equation}
A_{ij}=\begin{cases}\exp\!\big(-\|\mathbf{c}_i-\mathbf{c}_j\|^2/2\gamma^2\big)&\text{if }\|\mathbf{c}_i-\mathbf{c}_j\|\le\tau\\[2pt]0&\text{otherwise,}\end{cases}
\end{equation}

with kernel width $\gamma{=}2.0$\,cm and sparsity threshold $\tau{=}8.0$\,cm. However, to maintain a unified experimental protocol across all datasets and avoid dependence on dataset-specific electrode geometries, all experiments reported in this paper employ a fixed fully connected adjacency matrix as a neutral prior. This choice enables the GCN to learn functional relationships directly from the data while ensuring consistent graph construction across datasets.

The renormalized propagation operator is computed as
$\hat{\mathbf{A}}=\tilde{\mathbf{D}}^{-\frac{1}{2}}(\mathbf{A}+\mathbf{I}_C)\tilde{\mathbf{D}}^{-\frac{1}{2}}$,
where $\tilde{\mathbf{D}}_{ii}=\sum_j(\mathbf{A}+\mathbf{I}_C)_{ij}$ and the self-loops $\mathbf{I}_C$ preserve each electrode's intrinsic signal. Spatial aggregation yields

\[
\mathbf{H}_{\text{spatial}}
=
\operatorname{LayerNorm}
\!\left(
\sigma
\!\left(
\hat{\mathbf{A}}
\mathbf{X}_t
\mathbf{W}_{\text{gnn}}
\right)
\right)
\in
\mathbb{R}^{B\times C\times D},
\]

with learnable $\mathbf{W}_{\text{gnn}}\in\mathbb{R}^{L\times D}$ and ReLU activation $\sigma(\cdot)$. $\mathbf{H}_{\text{spatial}}$ serves as a spatial inductive bias for temporal modeling. The proposed graph formulation is flexible and can readily accommodate geometry-based, functional connectivity, or other anatomically informed adjacency matrices without modifying the network architecture. Thus, the geometry-informed adjacency presented above is a general formulation, whereas the experiments use a fixed fully connected adjacency to ensure consistency across datasets.

\subsection{Stage 2: Hierarchical Temporal Encoder--Decoder}
The U-Net encoder is a recurrence over $N{=}4$ downsampling levels, initialized by $\mathbf{H}^{(0)}=\mathbf{H}_{\text{spatial}}$. The diffusion timestep is encoded with sinusoidal embeddings $\mathbf{t}_{emb}$ projected by a SiLU-MLP and injected into each ResNet block as a learned bias shift, giving $\mathbf{S}^{(l)}=\mathbf{H}^{(l)}+\text{Conv1D}_{out}(\text{SiLU}(\text{GroupNorm}(\mathbf{h}^{(l)}_{res}+\mathbf{h}^{(l)}_{scale})))$, with $\mathbf{h}^{(l)}_{res}=\text{SiLU}(\text{GroupNorm}(\text{Conv1D}_{in}(\mathbf{H}^{(l)})))$ and $\mathbf{h}^{(l)}_{scale}=\text{MLP}(\text{SiLU}(\mathbf{t}_{emb}))$. Each $\mathbf{S}^{(l)}$ is stored as a skip connection and downsampled ($K{=}4$, stride~$2$); the compressed code $\mathbf{Z}=\mathbf{H}^{(N)}$ enters the bottleneck. The decoder mirrors the encoder: at each level the map is upsampled by transposed convolution, fused with the matching skip $\mathbf{D}^{(l)}_{fused}=\text{Concat}(\mathbf{D}^{(l)}_{\uparrow},\mathbf{S}^{(N-1-l)})$, and processed by a time-conditioned ResNet block, starting from $\mathbf{D}^{(0)}=\mathbf{Z}_{out}$ (Stage~3). The decomposition heads then map the output to $\hat{\mathbf{X}}_0=\mathbf{V}_{trend}+\mathbf{S}_{season}$.

\subsection{Stage 3: Mamba Bottleneck for Long-Range Temporal Modeling}
At the bottleneck, self-attention is replaced by a Bidirectional Mamba block, giving input-dependent long-range modeling with linear complexity $\mathcal{O}(L)$ versus the $\mathcal{O}(L^2)$ cost of Transformers. The underlying SSM has a diagonal state matrix $\mathbf{A}\in\mathbb{R}^{D\times N}$, input matrix $\mathbf{B}\in\mathbb{R}^{D\times N}$, and learns an input-dependent step size $\boldsymbol{\Delta}_k$ per position $k$:
\begin{equation}
\boldsymbol{\Delta}_k=\text{Softplus}(\text{Linear}(\mathbf{Z}_k)),
\end{equation}
\begin{equation}
\overline{\mathbf{A}}_k=\exp(\boldsymbol{\Delta}_k\odot\mathbf{A}),\quad
\overline{\mathbf{B}}_k=(\boldsymbol{\Delta}_k\odot\mathbf{A})^{-1}(\overline{\mathbf{A}}_k-\mathbf{I})\odot(\boldsymbol{\Delta}_k\odot\mathbf{B}),
\end{equation}
where $\odot$ is broadcast element-wise multiplication. The adaptive $\boldsymbol{\Delta}_k$ acts as a learned sampling rate: small values capture high-frequency transients, large values model slow background rhythms. We aggregate non-causal context with bidirectional scans $\mathbf{Z}_{mamba}=\text{Linear}(\text{Concat}(\mathbf{h}^{\rightarrow}(\mathbf{Z}),\mathbf{h}^{\leftarrow}(\mathbf{Z})))$ and decompose the latent into a polynomial trend $\mathbf{V}_{trend}=\mathcal{C}\cdot\text{Conv1D}(\text{GELU}(\text{Conv1D}(\mathbf{Z}_{mamba})))$ and a Fourier seasonality $\mathbf{S}_{season}=\mathcal{F}^{-1}(\text{TopK}(\mathcal{F}(\mathbf{Z}_{mamba})))$, with $\mathbf{Z}_{out}=\mathbf{V}_{trend}+\mathbf{S}_{season}$. This separates slow morphological drifts from fast oscillations as an inductive bias, but is trained end-to-end without strict stationarity or periodicity constraints.

\subsection{Stage 4: Latent-Space Guidance Classifier}
Guidance operates on the intermediate latent $\mathbf{h}^t$ rather than the heavily corrupted raw signal $\mathbf{X}_t$, keeping the classifier robust to early diffusion noise while enforcing pathology-relevant patterns. Given the noisy latent $\mathbf{X}_t\in\mathbb{R}^{B\times L\times C}$, we build multi-granular temporal tokens by partitioning the signal into overlapping patches $\{P_k\}_{k=1}^{G}$ ($G{=}3$, $P_1{=}16,P_2{=}32,P_3{=}64$) and projecting each into a shared $D$-dimensional space, $\mathbf{E}_t^{(k)}\in\mathbb{R}^{B\times N_k\times D}$, alongside channel-wise tokens $\mathbf{E}_c^{(m)}\in\mathbb{R}^{B\times C\times D}$ ($M{=}1$). To stabilize attention over long sequences we apply logarithmically-scaled rotary positional encoding with rotation angle
\begin{equation}
\Theta_j(\text{pos})=\text{base}^{-\frac{2\lfloor j/2\rfloor}{D}}\cdot\log(\text{pos}+1),\qquad \text{base}=10000,
\end{equation}
which moderates positional growth and improves numerical stability. The tokens are concatenated into $\mathbf{H}_0=\text{Concat}(\mathbf{E}_t^{(1)},\dots,\mathbf{E}_t^{(G)},\mathbf{E}_c^{(1)},\dots,\mathbf{E}_c^{(M)})\in\mathbb{R}^{B\times N\times D}$ ($N=\sum_k N_k+M\cdot C$) and processed by $L_{\text{enc}}{=}6$ time-conditioned Transformer layers, with $t$ injected through adaptive layer normalization; a global attention-pooling head yields $\mathbf{h}\in\mathbb{R}^{B\times K}$ over $K$ disease categories. During sampling, the noise prediction is modified as
\begin{equation}
\tilde{\epsilon}_t=\epsilon_\theta(\mathbf{h}^t,t)-s\sqrt{1-\bar{\alpha}_t}\,\nabla_{\mathbf{h}^t}\log p_\phi(y\mid\mathbf{h}^t),
\end{equation}
where $s\in[1,5]$ controls guidance strength and the factor $\sqrt{1-\bar{\alpha}_t}$ preserves stability. To preserve oscillatory structure we optimize a dual-domain loss combining time-domain reconstruction and frequency-domain spectral constraints,
\begin{equation}
\mathcal{L}=\|\mathbf{X}_0-\hat{\mathbf{X}}_0\|_1+\lambda\!\!\sum_{p\in\{\mathrm{Re},\mathrm{Im}\}}\!\!\|p(\mathcal{F}(\mathbf{X}_0))-p(\mathcal{F}(\hat{\mathbf{X}}_0))\|_1,
\end{equation}
with $\mathcal{F}$ the discrete Fourier transform and $\lambda{=}0.1$. Predicting $\mathbf{X}_0$ directly improves training stability and enables spectral preservation for physiologically realistic synthesis.


\section{Experiments and Results}

\subsection{Datasets and Implementation}
We evaluate on three public EEG datasets, all on a single NVIDIA RTX A4500 GPU (40\,GB). \textbf{ADSZ}~\cite{alves2022eeg}: 19-channel EEG at 128\,Hz, AD binary subset of 48 subjects (24 AD / 24 controls), 8--14\,s trials. \textbf{APAVA}~\cite{escudero2006analysis}: 16-channel EEG at 256\,Hz, 23 subjects (12 AD / 11 controls), 5\,s trials. \textbf{ADFD}~\cite{miltiadous2023dataset}: 19-channel, 500\,Hz, 88 subjects (36 AD, 23 FTD, 29 controls), long trials band-passed to 0.5--45\,Hz; excluding FTD yields \textbf{ADFD-Binary} (65 subjects, 53{,}182 samples). The generator uses Adam ($\beta_1{=}0.9,\beta_2{=}0.96$) over 25{,}000 epochs with EMA weights; the downstream classifier is trained separately for 3{,}000 epochs (Adam, lr~$1.0\times10^{-4}$).

\subsection{Reliability of Generated EEG for AD Diagnosis}
We use a subject-independent 80--20 split, train the diffusion model on the 80\% subset, and synthesize $\sim$1.5$\times$ extra samples. Classifiers are trained \emph{only} on synthetic data and evaluated on the unseen 20\% real test set, comparing ``Synthetic Train (TSTR)'' against ``Real Train (TRTR)'' on identical held-out data (Table~\ref{tab:baseline_accuracy}). Models trained purely on synthetic AD data match or exceed the real-data baseline: ADFormer attains 97.14\% on ADSZ under synthetic training, surpassing its real-data result (93.08\%) and all competing architectures, and reaches 51.22\% on the harder ADFD 3-Class task---indicating the generated signals retain diagnostically meaningful patterns that generalize to unseen subjects.

\begin{table}[t]
\centering
\caption{Classification accuracy (\%) trained on our generated data (Synthetic, TSTR) versus the original dataset (Real, TRTR), tested on identical held-out real data. A representative set of baselines is shown.}
\label{tab:baseline_accuracy}
\resizebox{0.92\linewidth}{!}{
\begin{tabular}{lrrrrrrrr}
\toprule
 & \multicolumn{2}{c}{\textbf{ADSZ}} & \multicolumn{2}{c}{\textbf{APAVA}} & \multicolumn{2}{c}{\textbf{ADFD Binary}} & \multicolumn{2}{c}{\textbf{ADFD 3 Class}} \\
\cmidrule(lr){2-3} \cmidrule(lr){4-5} \cmidrule(lr){6-7} \cmidrule(lr){8-9}
\textbf{Model} & \textbf{Syn} & \textbf{Real} & \textbf{Syn} & \textbf{Real} & \textbf{Syn} & \textbf{Real} & \textbf{Syn} & \textbf{Real} \\
\midrule
ADFormer \cite{wang2024adformer} & \textbf{97.14} & 93.08 & \textbf{82.81} & 80.00 & \textbf{73.74} & 75.80 & \textbf{51.22} & 55.21 \\
Autoformer \cite{wu2021autoformer} & {93.61} & 91.43 & 69.89 & 68.64 & 61.31 & 59.81 & 44.28 & 45.25 \\
Informer \cite{zhou2021informer} & 93.18 & 90.22 & 74.86 & 73.11 & 59.95 & 61.88 & 46.90 & 48.45 \\
ITransformer \cite{liu2023itransformer} & 78.91 & 78.79 & 75.01 & 74.55 & 71.44 & 73.85 & 49.97 & 52.60 \\
PatchTST \cite{nie2022time} & 79.17 & 79.12 & 69.86 & 67.03 & 60.29 & 59.10 & 44.29 & 44.37 \\
Pyraformer \cite{liu2021pyraformer} & 91.89 & 90.77 & 81.20 & 79.54 & 65.10 & 66.00 & 49.27 & 51.52 \\
TimesNet \cite{wu2022timesnet} & 92.59 & 92.75 & 76.31 & 76.30 & 60.27 & 61.79 & 47.14 & 48.81 \\
EEGNet \cite{lawhern2018eegnet} & 89.24 & 86.81 & 71.22 & 71.81 & 65.28 & 71.91 & 47.26 & 51.78 \\
TCN \cite{bai2018empirical} & 82.18 & 88.90 & 80.29 & 84.71 & 59.21 & 68.71 & 42.38 & 50.17 \\
\bottomrule
\end{tabular}}
\end{table}

\subsection{Comparison with Generative Peer Models}
We compare synthesis quality against TimeGAN~\cite{yoon2019time}, TimeVAE~\cite{desai2021timevae}, and Diffusion-TS~\cite{yuan2024diffusion}, holding the downstream classifier strictly constant so differences reflect generative fidelity alone (Table~\ref{tab:peer_comparison}). As TimeGAN and TimeVAE are single-class generators, we train one model per class and merge outputs for multi-class data. GraM-Diff outperforms all baselines across Accuracy, Precision, F1, AUROC, and AUPRC. On ADSZ it reaches 97.18\% accuracy, far above Diffusion-TS (87.41\%) under identical conditions and the segregated TimeGAN (69.81\%) and TimeVAE (64.25\%).

\begin{table}[t]
\centering
\caption{Comparison with state-of-the-art generative peer models across ADSZ, APAVA, and ADFD (Binary). ROC and PRC denote AUROC and AUPRC.}
\label{tab:peer_comparison}
\resizebox{\textwidth}{!}{

\begin{tabular}{lrrrrrrrrrrrrrrr}
\toprule
\multirow{2}{*}{\textbf{Model}} & \multicolumn{5}{c}{\textbf{ADSZ}} & \multicolumn{5}{c}{\textbf{APAVA}} & \multicolumn{5}{c}{\textbf{ADFD (Binary)}} \\
\cmidrule(lr){2-6} \cmidrule(lr){7-11} \cmidrule(lr){12-16}
& \textbf{Acc} & \textbf{Prec} & \textbf{F1} & \textbf{ROC} & \textbf{PRC} & \textbf{Acc} & \textbf{Prec} & \textbf{F1} & \textbf{ROC} & \textbf{PRC} & \textbf{Acc} & \textbf{Prec} & \textbf{F1} & \textbf{ROC} & \textbf{PRC} \\
\midrule
TimeGAN \cite{yoon2019time} & 69.81 & 70.28 & 71.93 & 72.29 & 69.35 & 54.28 & 58.19 & 58.92 & 59.70 & 59.00 & 51.35 & 52.63 & 52.74 & 51.70 & 54.36 \\
TimeVAE \cite{desai2021timevae} & 64.25 & 68.11 & 69.74 & 65.29 & 67.81 & 53.91 & 52.14 & 52.83 & 58.51 & 55.92 & 52.10 & 51.88 & 53.83 & 58.14 & 59.53 \\
Diffusion-TS \cite{yuan2024diffusion} & 87.41 & 89.25 & 88.65 & 90.83 & 90.24 & 63.87 & 64.99 & 63.81 & 70.96 & 72.39 & 55.39 & 58.31 & 57.12 & 60.25 & 65.41 \\
\textbf{Ours} & \textbf{97.18} & \textbf{97.20} & \textbf{96.28} & \textbf{97.82} & \textbf{95.20} & \textbf{82.81} & \textbf{85.99} & \textbf{84.21} & \textbf{89.85} & \textbf{88.46} & \textbf{73.74} & \textbf{75.93} & \textbf{77.31} & \textbf{77.28} & \textbf{79.21} \\
\bottomrule
\end{tabular}}
\end{table}

\subsection{Generative Fidelity}
We assess fidelity with Context-FID (distributional distance in deep feature space) and Correlation Score (preservation of spatio-temporal dependencies), lower being better, against Diffusion-TS. GraM-Diff attains markedly lower Context-FID on every dataset (ADSZ 0.109 vs.\ 0.585; APAVA 0.286 vs.\ 0.426; ADFD-Binary 0.484 vs.\ 1.030; ADFD 3-Class 0.529 vs.\ 0.920) and tighter Correlation Scores (ADSZ 0.275 vs.\ 0.492; APAVA 0.402 vs.\ 0.829; ADFD-Binary 0.532 vs.\ 2.890; ADFD 3-Class 0.583 vs.\ 2.120), confirming closer agreement with real EEG dynamics.

\subsection{Downstream Utility, Ablation, and Sensitivity}
\textbf{Utility and augmentation.} Beyond statistical similarity, we test imputation (masking 10--90\%), forecasting (horizons 4--32), and real/synthetic mixture classification. Errors stay low under heavy difficulty: on ADSZ, imputation MSE rises only to 0.108 at 90\% masking and forecasting MSE remains 0.0501 at horizon~32. Augmentation accuracy is preserved even in heavily synthetic regimes (96.87\% on ADSZ at 80\% synthetic), confirming the synthetic data's value for limited clinical datasets.

\textbf{Ablation.} Removing components (Table~\ref{tab:ablation}, top) degrades performance, with Patch Embeddings + LogRoPE most critical: accuracy drops from 97.18\% to 91.81\% on ADSZ and from 82.81\% to 69.65\% on APAVA, underscoring its role in long-range temporal and positional modeling. The GNN and Mamba modules each contribute additional gains.

\textbf{Sensitivity.} Injecting Gaussian noise (Table~\ref{tab:ablation}, bottom) degrades performance gracefully: on ADSZ the model retains an AUROC of 85.35\% under 10\% noise (vs.\ 97.82\% baseline), with a similar trend on APAVA, indicating robustness to moderate signal perturbation.

\begin{table}[t]
\centering
\caption{Ablation (top) and Gaussian-noise sensitivity (bottom) on ADSZ and APAVA.}
\label{tab:ablation}
\resizebox{0.82\linewidth}{!}{

\begin{tabular}{lrrrrr}
\toprule
\multirow{2}{*}{\textbf{Setup}} & \multicolumn{3}{c}{\textbf{ADSZ}} & \multicolumn{2}{c}{\textbf{APAVA}} \\
\cmidrule(lr){2-4}\cmidrule(lr){5-6}
& \textbf{Acc} & \textbf{F1} & \textbf{AUROC} & \textbf{Acc} & \textbf{AUROC} \\
\midrule
Full Model & \textbf{97.18} & \textbf{96.28} & \textbf{97.82} & \textbf{82.81} & \textbf{89.85} \\
w/o GNN & 95.80 & 94.99 & 96.82 & 79.31 & 84.83 \\
w/o Mamba & 95.23 & 95.28 & 95.27 & 78.45 & 82.36 \\
w/o Patch+LogRoPE & 91.81 & 92.84 & 95.43 & 69.65 & 67.52 \\
\midrule
\multicolumn{6}{l}{\textit{Gaussian noise sensitivity}} \\
0\% (Base) & \textbf{97.18} & \textbf{96.28} & \textbf{97.82} & \textbf{82.81} & \textbf{89.85} \\
10\% & 79.20 & 81.54 & 85.35 & 71.77 & 78.42 \\
20\% & 72.98 & 77.31 & 75.30 & 62.85 & 71.20 \\
\bottomrule
\end{tabular}}
\end{table}

\section{Conclusion}
We presented GraM-Diff, a unified classifier-guided Graph--Mamba diffusion framework coupling topology-aware spatial priors with linear-complexity long-range temporal modeling for conditional multichannel EEG synthesis. A single model generates both healthy and pathological signals within a shared latent space, producing high-fidelity samples that better match real distributions and improve downstream AD classification, with superior Context-FID and correlation scores and graceful robustness under noise. Future work will extend the framework to multimodal and irregularly sampled time series while improving scalability and controllability for clinical deployment.

\vspace{2mm}

\noindent\textbf{Acknowledgments.} The work of Md Sajid is supported by R\&D Section, Indian Institute of Technology Indore, under the Translation Research Fellowship (TRF). This work is supported by PraxiaTech Private Limited. Further, this work is supported by IITI DRISHTI CPS Foundation under the National Mission on Interdisciplinary Cyber Physical System (NM-ICPS) of the Department of Science and Technology, Government of India.

\vspace{2mm}

\noindent\textbf{Disclosure of Interests.} The authors have no competing interests.

\bibliographystyle{splncs04}
\bibliography{refs}

\end{document}